\documentclass[letterpaper]{article} 
\usepackage[preprint]{aaai2027} 
\usepackage[hyphens]{url} 
\usepackage{graphicx} 
\usepackage{natbib} 
\usepackage{caption} 
\usepackage{amsmath,amssymb}
\usepackage{booktabs}
\usepackage{multirow}
\usepackage{tabularx}
\title{UrbanAgent: A Tool-Augmented Agent for Cross-System Urban Tasks}
\author{Jiayu~Cao,~
        Xingyuan~Zeng,~
        Feiyu~Li,~
        Zhijin~Huang,~
        Xujie~Yuan,~
        Rongxiang~Chen,~
        Shimin~Di,~
        Libin~Zheng,~
        and~Jian~Yin}
\affiliations{}

\begin{document}
\maketitle

\begin{abstract}
Modern cities rely on an increasing number of digital services to operate, but residents’ daily needs are still difficult to meet. Services are fragmented and have little interoperability, placing a heavy operational burden on users. Existing digital platforms, urban foundation models, and intelligent assistants each address only isolated aspects of an urban task. But they struggle to reliably convert complex natural-language requests into executable cross-system workflows. We propose UrbanAgent, a tool-augmented agent framework for cross-system urban tasks. It couples the cognitive and reasoning capabilities of a large language model with a toolset supporting code execution, API calls, and Model Context Protocol. Through one adaptive closed loop, it clarifies missing information before acting, grounds tool use in live observations, and aligns the final response with observed evidence and task constraints. To address the evaluation gap, we introduce UrbanEval, a benchmark specifically designed for cross-system urban request. Unlike prior benchmarks that assess either general tool use or urban knowledge and reasoning, UrbanEval evaluates both task results and execution quality, including required tool coverage, dependency validity, and evidence traceability. Experimental results indicate that UrbanAgent reaches a 71\% task success rate, 10points above the strongest baseline. This lead holds across GPT-5-mini, Gemini-2.5-flash, DeepSeek-V4-flash, and Qwen3-235B-A22B.
\end{abstract}


\section{Introduction}
\begin{figure*}[t]
    \centering
    \includegraphics[width=0.9\textwidth]{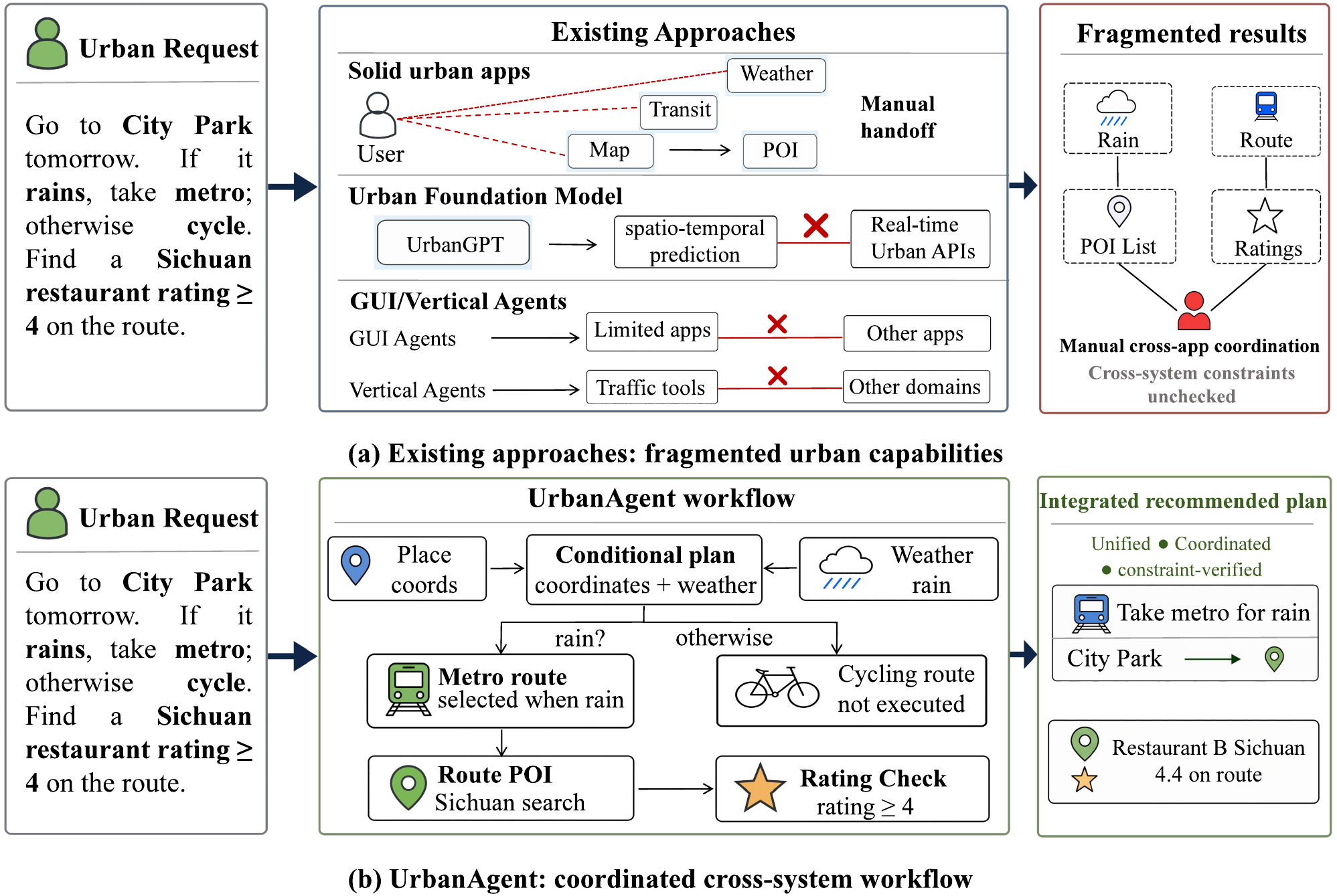}
    \caption{
    Fragmented urban services vs UrbanAgent. (a)Existing approaches each cover one fragment of a compositional request. (b) UrbanAgent executes the same request as one coordinated, evidence-aligned cross-system workflow.}
    \label{fig:intro}
\end{figure*}

Modern cities have deployed extensive digital systems for transportation, urban planning, environment, energy, and public services~\cite{zheng2014urban}.Yet this digitalization has not translated into convenience. These functions are scattered across separate systems with independent data, interfaces, and procedures. Even a simple daily request involves multiple aspects. Figure~\ref{fig:intro} illustrates a representative request that combines weather-related route planning with restrictions on restaurants along the way. Existing services can satisfy individual subtasks, but the user must manually transfer intermediate results, compare alternatives, and verify the final combination. Such requests are typically multi hop and compositional, spanning multiple systems, locations, and constraints~\cite{he2025localsearchbench}. The manual coordination required is especially burdensome for older residents, who need more time to find the right function in feature-rich interfaces and make more errors~\cite{yu2024reducing}.

Existing research addresses separate components of the coordination problem. Integrated urban platforms such as Hangzhou City Brain aggregate municipal data and services into a unified operating layer for traffic, public safety, and environmental management~\cite{zhang2019citybrain}. However each deployment remains bounded by a single domain or operating organization and does not act on an individual resident's cross-system request. Urban foundation models such as UrbanGPT and CityGPT support spatiotemporal prediction and spatial reasoning, but do not interact with real-time service systems~\cite{li2024urbangpt,feng2024citygpt}. Agent systems go a step further and take actions on real applications and services, yet each agent is confined either to a single application's interface or to a single predefined task. GUI agents such as AppAgent and Mobile-Agent focus on interactions within a single application~\cite{zhang2025appagent,wang2024mobileagent}, while urban vertical agents such as UrbanKGent and LLMLight execute specific domain tasks, including knowledge graph construction and traffic signal control~\cite{ning2024urbankgent,lai2024llmlight}. Existing approaches integrate urban data, reason about cities, or act within a single application or domain, but do not coordinate heterogeneous services around a complete user request.

Tool-augmented LLM agents can coordinate independent services through standardized tool interfaces. They select and invoke tools, then pass intermediate results between them. Prior work has applied such agents to web navigation and software engineering~\cite{yao2022react,qin2024toolllm,zhou2024webarena,xie2024osworld,jimenez2024swebench}. These advances make general tool agents a promising basis for executing cross-system urban tasks. However, direct transfer is not feasible. Urban requests are usually expressed as high-level goals rather than complete execution paradigms. Fulfilling them may require real-time environmental conditions, spatial relations, transportation options, and local-service information from providers with different geographic coverage. The workflow also emerges during execution because later calls may depend on places or conditions returned earlier. Although none of these challenges is unique in isolation, their combination makes urban workflows particularly fail. When a user asks for the “nearest” service without providing a location, an agent may silently assume one. Then it can retrieve a valid place, obtain current weather, and compute an accurate route for the wrong location. Each tool call may appear successful, but their combination produces a plausible but unusable recommendation. Therefore, reliable execution requires clarification before action, consistency checks, ordered tool use, and constraint verification.

To bridge this gap, we propose UrbanAgent, a tool-augmented agent framework for open urban tasks. It owns four components including cognition, a reasoning--execution core, tool invocation, and synthesis. They clarify missing critical task inputs before taking action, select each operation based on previous observations, rely on call consumption of upstream values, recover from empty or geo-invalid returns, and align the final response with collected evidence and task constraints. Every action and statement is grounded in tool observations rather than the model's prior knowledge.

Furthermore, we find no benchmark evaluates agents on coordinating heterogeneous urban services across domains. General tool-use benchmarks test function calls in non-urban settings~\cite{patil2025bfcl,qin2024toolllm,xu2023agentbench}. Urban benchmarks test city knowledge, reasoning, or simulation, not live service execution~\cite{feng2025citybench,zhou2025urbench}. Agentic local life search stays within a single domain category~\cite{he2025localsearchbench}. To bridge this gap, we construct UrbanEval. The requests range from ambiguous tasks that require clarification to constrained tasks with ordered dependencies. Because a plausible answer can conceal an incomplete or invalid execution trace, UrbanEval assesses both task results and execution quality, including tool coverage, dependency order, and evidence grounding.

\begin{figure*}[t]
    \centering
    \includegraphics[width=\textwidth]{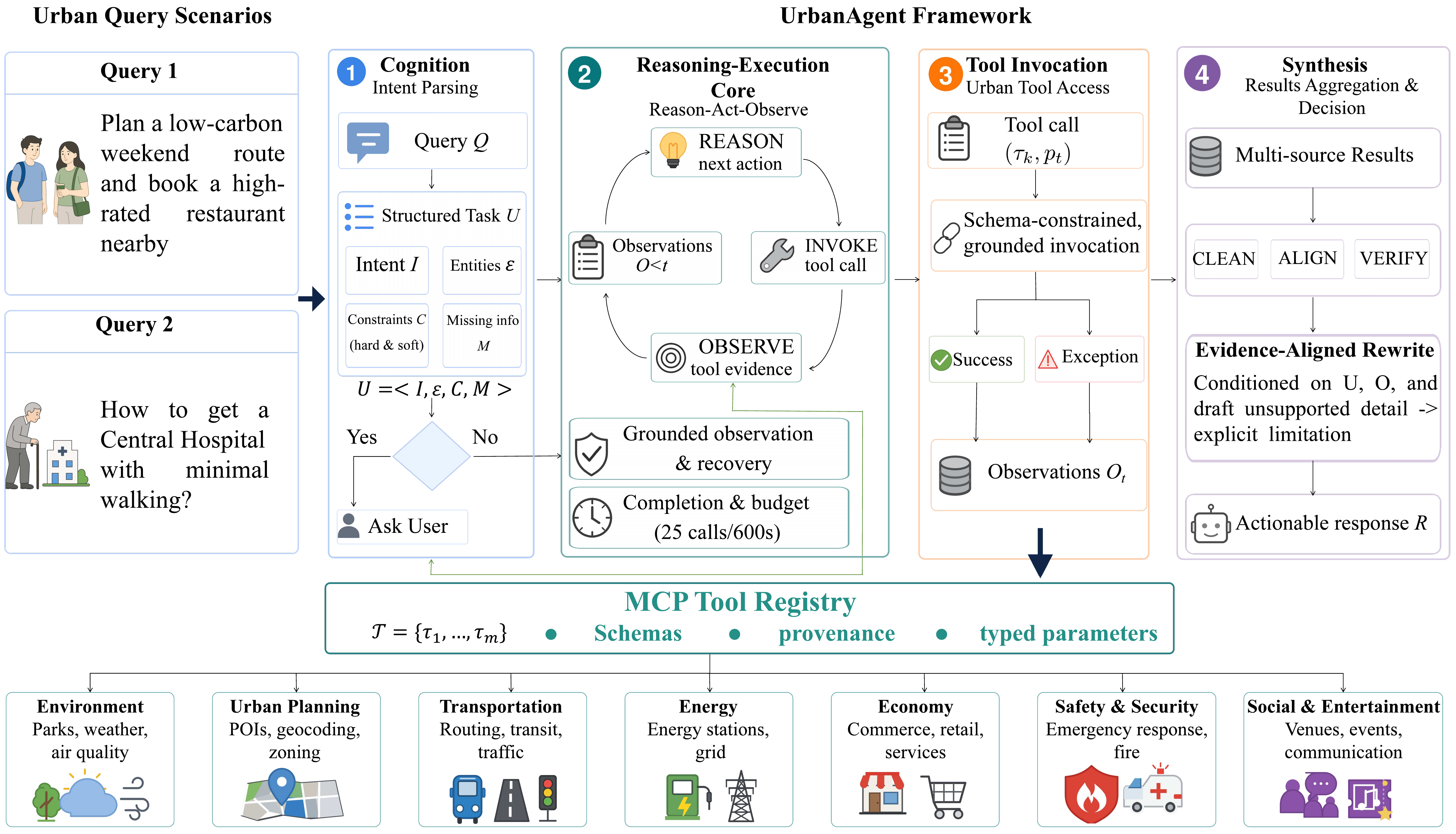}
    \caption{Framework of UrbanAgent. (1)~Cognition maps a query to a structured task $\mathcal{U}$ and requests clarification when a key task parameter is missing. (2)~The reasoning--execution core selects the next action from prior observations, and (3)~tool invocation executes that action through the unified registry $\mathcal{T}$; the two alternate until the agent commits a preliminary answer $\hat{\mathcal{R}}$. (4)~Synthesis aligns the draft with multi-source observations, verifies task constraints, and commits the final answer $\mathcal{R}$.}
    \label{fig:framework}
\end{figure*}

To address these gaps, we make the following contributions:
\begin{itemize}
    \item We propose UrbanAgent to bridge the gap between high-level user requests and fragmented urban services. It clarifies missing parameters, preserves cross-service dependencies, and grounds the final response in validated tool observations.
    
    \item We introduce UrbanEval to make cross-system urban execution evaluable. It pairs 250 requests with trace-aware criteria that verify missing handling, required capability coverage, dependency preservation, and evidence support. Therefore, it distinguishes between truly completed workflows and plausible answers produced through incomplete execution.
    
    \item We show that UrbanAgent improves cross-system task completion over general orchestration baselines under matched models and tools. With GPT-5-mini, UrbanAgent completes 64.0\% of executable tasks vs 53.5\% for the strongest baseline, and it achieves the highest TSR across all tested base models.
\end{itemize}

\section{UrbanAgent}

\subsection{Problem Formulation and Overview}

Let $\mathcal{Q}$ denote a natural-language query and $\mathcal{T}=\{\tau_1,\ldots,\tau_m\}$ a registry of urban tools. Each tool records its name, description, source server, and input JSON Schema. UrbanAgent accesses the external urban state $\mathcal{W}_t$ through tool calls: invoking $\tau_k$ with arguments $p_t$ yields an observation $o_t$. The agent combines $\mathcal{Q}$ with these observations to produce $\mathcal{R}=\Phi(\mathcal{Q},\mathcal{W}_t;\mathcal{T})$~\cite{yao2022react,qin2024toolllm,mcp2025spec}.

UrbanAgent addresses four recurring failures of cross-system urban tasks through the components in Figure~\ref{fig:framework}. Cognition detects missing inputs (e.g., location or time) and clarifies before acting. The reasoning--execution core selects actions incrementally so downstream calls consume upstream values. Tool invocation validates service returns and recovers from failed or geographically inconsistent calls. And synthesis checks the draft against task constraints and collected evidence. We evaluate this integrated design rather than attributing gains to any single component.

\subsection{Cognition and Clarification}

Cognition maps the query to a structured task,
$f_{\text{cog}}^{\pi}:\mathcal{Q}\mapsto\mathcal{U}$, where
$\mathcal{U}=\langle\mathcal{I},\mathcal{E},\mathcal{C}_{\text{hard}},\mathcal{C}_{\text{soft}},\mathcal{M}\rangle$.
The intent $\mathcal{I}$ specifies the task objective, and $\mathcal{E}$ records the entities and parameters required for subsequent tool calls. $\mathcal{C}_{\text{hard}}$ encodes mandatory conditions and $\mathcal{C}_{\text{soft}}$ ranking preferences.

The set $\mathcal{M}$ collects missing user-specific inputs that cannot be inferred without changing the task. If $\mathcal{M}\neq\varnothing$, UrbanAgent requests these inputs and terminates before any tool call, rather than inferring a location, destination, or other task-defining value~\cite{li2025urbanllm,qin2024toolllm}.

\subsection{Reasoning--Execution Core}

The reasoning--execution core implements a reason--act--observe policy over the structured task $\mathcal{U}$ and observation history $\mathcal{O}_{<t}$. It selects actions incrementally rather than constructing a complete tool sequence in advance. At turn $t$, $\pi$ selects either a tool $\tau_k$ with arguments $p_t$ or the terminal action $\operatorname{Commit}(\hat{\mathcal{R}})$:
\begin{equation}
\begin{aligned}
a_t &= \pi\!\left(\mathcal{U},\mathcal{O}_{<t}\mid
\operatorname{Schema}(\mathcal{T})\right), \\
a_t &\in \{(\tau_k,p_t),\operatorname{Commit}(\hat{\mathcal{R}})\}.
\end{aligned}
\end{equation}
For $a_t=(\tau_k,p_t)$, the runtime executes the call and appends the result, $\mathcal{O}_{\leq t}=\mathcal{O}_{<t}\cup\{o_t\}$; selecting the next action from this history lets a downstream call consume upstream values. Execution ends when $a_t=\operatorname{Commit}(\hat{\mathcal{R}})$.


\begin{figure*}[t]
    \centering
    \includegraphics[width=\textwidth]{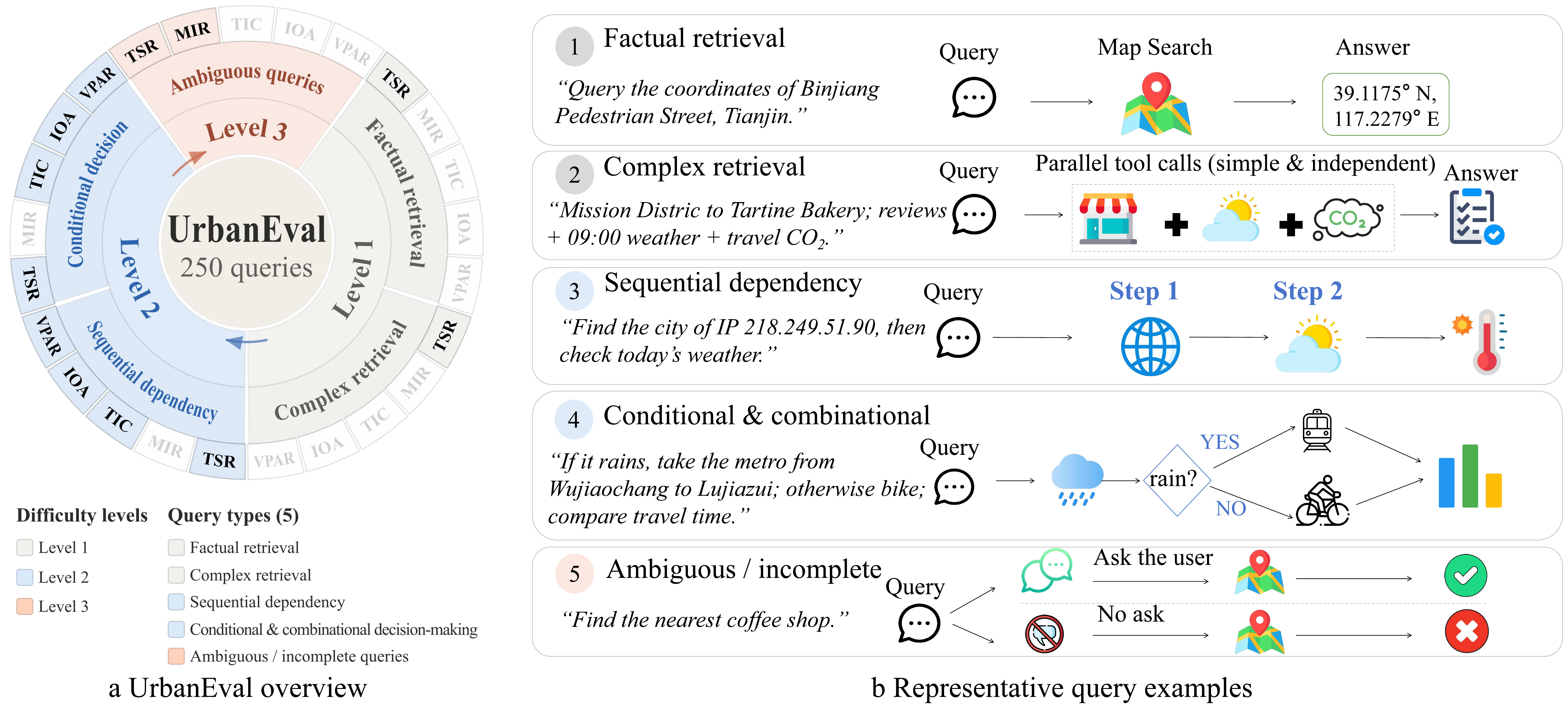}
    \caption{Overview of UrbanEval. (a)~The 250 queries span five task categories and three difficulty levels, and every category is scored by result-layer metrics, and the Level-2 category additionally by process-layer metrics. (b)~Representative query examples and tool workflows for the five categories.}
    \label{fig:dataset_evaluation}
\end{figure*}

\subsection{Tool Invocation and Grounding}

At each turn, the function-calling interface exposes the registered tool names, descriptions, and input schemas to $\pi$. These schemas constrain the action space and the shape of $p_t$ but cannot ensure semantic validity. The runtime converts unknown tool names, malformed arguments, or backend exceptions into error observations, from which $\pi$ revises the arguments or switches tools on the next turn~\cite{mcp2025spec,patil2025bfcl}. Each observation $o_t$ records the tool name, source server, parameters, raw return, status, and latency. The runtime caches a call signature only after it returns usable data and rejects an exact repeat, while still allowing failed calls to be retried. After eight consecutive failures on the same tool, it advises switching tools or reporting the subtask as unavailable.

Urban services may also return syntactically valid but unusable results. A geographic-consistency check rejects a geocode whose city or province conflicts with the requested region, and an empty-result check rejects a return without usable content. Both set the observation status to failed and report the failure to $\pi$. Policy instructions further encode provider coverage and call prerequisites, so $\pi$ avoids mainland-only geocoders for overseas locations and resolves POI coordinates before route planning. All observations remain in $\mathcal{O}$, but only valid returns support factual claims.

\subsection{Evidence-Aligned Synthesis}
The reasoning--execution core returns a draft $\hat{\mathcal{R}}$, which synthesis refines conditioned on the structured task and observations:
\begin{equation}
\mathcal{R}=f_{\text{syn}}^{\pi}
(\mathcal{U},\mathcal{O},\hat{\mathcal{R}}).
\end{equation}
A draft may narrate the execution process, defer requested actions to a follow-up turn, or overlook the most specific observation. Synthesis corrects these through the three operations in Figure~\ref{fig:framework}: \emph{Clean} removes process descriptions and redundant follow-up offers; \emph{Align} selects the most relevant evidence in $\mathcal{O}$ and matches it to the requested format; \emph{Verify} checks the conclusion against the constraints in $\mathcal{U}$ and admits no fact absent from $\mathcal{O}$ and $\hat{\mathcal{R}}$. When observations lack the needed detail, $\mathcal{R}$ states the limitation rather than inferring a value. If synthesis fails or returns empty, the runtime falls back to $\hat{\mathcal{R}}$.

\subsection{Implementation Details}
UrbanAgent requires no task-specific training. The evaluated configuration uses the ReAct core with evidence-aligned rewrite synthesis, and remote MCP services and local computational tools share one registry interface~\cite{anthropic2024mcp,mcp2025spec,ouyang2025code2mcp}. The controlled experiments use GPT-5-mini at temperature $T=1.0$ with a per-query budget of 25 LLM calls or 600 seconds. When at most two calls or 90 seconds remain, the runtime disables tool calls and requests a draft from the available observations.

\section{UrbanEval Benchmark}

\begin{table}[t]
\centering
\setlength{\tabcolsep}{3pt}
{\small
\begin{tabularx}{\columnwidth}{@{}l l X@{}}
\toprule
Layer & Metric & Definition \\
\midrule
\multirow{2}{*}{Result} & TSR & Task success rate --- a correct, complete, constraint-satisfying answer \\[2pt]
 & MIR & Missing-information identification rate --- guess-free flagging of a missing task-critical input \\
\midrule
\multirow{3}{*}{Process} & TIC & Tool invocation completeness --- coverage of the required tool capabilities \\[2pt]
 & IOA & Invocation order accuracy --- dependency pairs called in the required order \\[2pt]
 & VPAR & Valid proactive augmentation rate --- useful augmentation or verification beyond the minimum \\
\midrule
\multirow{2}{*}{Efficiency} & Tokens per query & Mean over all queries \\[2pt]
 & Tokens per success & Mean over successful tasks only \\
\bottomrule
\end{tabularx}
}
\caption{UrbanEval’s three-layer metric mechanism. TIC, IOA, and the efficiency measures are computed deterministically from recorded traces; TSR, MIR, and VPAR are scored by an LLM judge (Qwen3-Max) under a shared rubric. Process metrics are reported on Level 2, and VPAR only on runs having at least one tool call.}
\label{tab:metrics}
\end{table}

\noindent Existing tool-use benchmarks are not designed around cross system urban workflows~\cite{patil2025bfcl,qin2024toolllm,xu2023agentbench}, while urban benchmarks focus on scene understanding or simulation rather than live service execution~\cite{feng2025citybench,zhou2025urbench}. UrbanEval instead tests high level requests that combine missing inputs, service dependencies, time-varying observations, and evidence-grounded constraints.

\begin{table*}[!t]
\centering
\setlength{\tabcolsep}{4pt}
{\small
\begin{tabular*}{\textwidth}{@{\extracolsep{\fill}}l l *{7}{c}@{}}
\toprule
 & & \multicolumn{2}{c}{Results} & \multicolumn{2}{c}{Efficiency} & \multicolumn{3}{c}{Process} \\
\cmidrule(lr){3-4}\cmidrule(lr){5-6}\cmidrule(lr){7-9}
Base model & Method & TSR~$\uparrow$ & MIR~$\uparrow$ & Tok.~$\downarrow$ & T/Succ.~$\downarrow$ & TIC~$\uparrow$ & IOA~$\uparrow$ & VPAR~$\uparrow$ \\
\midrule
\multirow{5}{*}{\shortstack[l]{GPT-5-mini\\(main)}}
 & Native tool-calling & \underline{0.61} & 0.92 & 26.7 & 42.8 & 0.83 & \underline{0.99} & 0.83 \\
 & ReAct & 0.57 & 0.94 & \textbf{21.8} & \textbf{37.3} & 0.75 & 0.96 & \underline{0.84} \\
 & Plan-and-Execute & 0.47 & 0.90 & 133.1 & 272.8 & 0.79 & 0.96 & \underline{0.84} \\
 & AutoGen & \underline{0.61} & \underline{0.96} & \underline{24.5} & \underline{40.1} & \underline{0.86} & 0.98 & 0.83 \\
 & \textbf{UrbanAgent} & \textbf{0.71} & \textbf{0.98} & 83.8 & 118.3 & \textbf{0.89} & \textbf{1.00} & \textbf{0.96} \\
\midrule
\multirow{5}{*}{Gemini-2.5-flash}
 & Native tool-calling & 0.37 & 0.62 & 78.1 & 210.0 & 0.85 & \textbf{1.00} & 0.79 \\
 & ReAct & 0.37 & 0.70 & \textbf{46.2} & \underline{125.5} & 0.86 & \textbf{1.00} & 0.82 \\
 & Plan-and-Execute & \underline{0.39} & \underline{0.72} & 79.4 & 202.6 & \textbf{0.92} & 0.94 & \underline{0.90} \\
 & AutoGen & 0.31 & 0.18 & 89.3 & 286.2 & \underline{0.87} & \underline{0.99} & \textbf{0.93} \\
 & \textbf{UrbanAgent} & \textbf{0.53} & \textbf{0.98} & \underline{52.9} & \textbf{99.4} & 0.85 & 0.98 & 0.86 \\
\midrule
\multirow{5}{*}{DeepSeek-V4-flash}
 & Native tool-calling & \underline{0.50} & \underline{0.96} & \underline{64.0} & \underline{129.0} & \underline{0.92} & 0.97 & \underline{0.96} \\
 & ReAct & 0.48 & \underline{0.96} & \textbf{58.7} & \textbf{122.3} & \textbf{0.95} & \textbf{0.99} & 0.95 \\
 & Plan-and-Execute & 0.42 & 0.82 & 112.9 & 268.8 & 0.88 & 0.97 & \textbf{0.98} \\
 & AutoGen & 0.21 & 0.20 & 107.8 & 508.6 & 0.91 & \underline{0.98} & 0.92 \\
 & \textbf{UrbanAgent} & \textbf{0.52} & \textbf{0.98} & 104.3 & 199.0 & 0.85 & \textbf{0.99} & 0.94 \\
\midrule
\multirow{5}{*}{Qwen3-235B-A22B}
 & Native tool-calling & 0.30 & 0.82 & \textbf{26.2} & \textbf{88.6} & 0.69 & \underline{0.82} & 0.46 \\
 & ReAct & 0.28 & \underline{0.86} & \underline{31.4} & 113.9 & 0.68 & 0.76 & 0.53 \\
 & Plan-and-Execute & \underline{0.32} & 0.84 & 83.3 & 263.5 & \underline{0.74} & 0.70 & \underline{0.72} \\
 & AutoGen & 0.08 & 0.06 & 128.6 & 1607.8 & \textbf{0.78} & 0.74 & 0.62 \\
 & \textbf{UrbanAgent} & \textbf{0.41} & \textbf{0.96} & 40.4 & \underline{98.9} & 0.68 & \textbf{0.95} & \textbf{0.86} \\
\bottomrule
\end{tabular*}
}
\caption{Main results on UrbanEval for tool-augmented systems, grouped by base model. GPT-5-mini defines the controlled main setting; the other base models evaluate generalization under the same tool set, budget, and protocol.}
\label{tab:main_results}
\end{table*}

UrbanEval addresses this gap with 250 natural-language requests executed against external services exposed through MCP servers across seven urban-computing domains~\cite{zheng2014urban}. Its challenge structure varies the form of execution rather than merely increasing the number of tool calls. As shown in Figure~\ref{fig:dataset_evaluation}, five query types cover factual retrieval, complex retrieval across independent services, sequential dependency, conditional or combinatorial decision-making, and ambiguous or incomplete intent. These types are organized into three difficulty levels. Level 1 tests information acquisition and composition. Level 2 tests dependency-aware execution and constrained decision-making. Level 3 tests whether an agent identifies task-critical omissions before acting. Each type contains 50 queries, providing balanced coverage of retrieval, coordination, branching, and clarification failures.

A fluent answer does not prove the task was carried out: an agent can skip a required tool, call tools in the wrong order, or state a fact, such as a travel time, that no tool ever returned. UrbanEval therefore scores each request against a human-written checklist of small, independently verifiable criteria instead of one gold answer. Each criterion states what must hold (e.g., the returned hospital is currently open), which tool capability and upstream result it depends on (a place search must run before, and feed, the route query), and any user input the agent must clarify first. Evidence is matched by type. Specifically, fixed facts such as a district's official name are checked against fixed references. Facts that change over time such as current weather are compared with the raw tool outputs from. And computed values such as an arrival estimate are compared with the specific observations. Therefore, the value is only recognized when the trajectory actually generates the input behind it.

The schema drives a three-layer metric mechanism (Table~\ref{tab:metrics}) that separates three questions answer-only evaluation conflates: was the request solved, was the execution valid, and what did success cost? Process metrics are reported on Level~2, where explicit tool dependencies are defined, and VPAR only on runs that issue at least one tool call. Together, these layers distinguish plausible answers from grounded task completion, valid orchestration, and execution cost.

\section{Experiments}
\noindent Our experiments test one mechanism: clarifying missing inputs before acting, then executing under explicit tool dependencies and task constraints. We ask whether this structure completes the complex urban tasks that baseline agents leave unfinished, under matched model and tool access. Three questions organize the study. \textbf{Q1 (Effectiveness and cost):} Does UrbanAgent improve end-to-end task success, and at what token cost? \textbf{Q2 (Process):} Which execution behaviors, tool coverage, dependency order, augmentation, and clarification, accompany its gains? \textbf{Q3 (Robustness):} Does the advantage persist across closed and open base models?

\subsection{Experimental Setup}
\noindent\textbf{Baselines.} We evaluate four tool-augmented baselines that represent established orchestration strategies: Native tool-calling, which delegates action selection to the model's function-calling interface; ReAct, which interleaves reasoning and tool use~\cite{yao2022react}; Plan-and-Execute, which plans before execution~\cite{wang2023plan}; and AutoGen, which pairs an assistant with a tool executor~\cite{wu2023autogen}. To compare orchestration rather than access, every tool-augmented system within a base-model block receives the same queries, MCP tools, tool descriptions, JSON schemas, temperature, and budget (25 LLM calls or 600 seconds per query). Because clarification is one of the behaviors we measure, every system may ask the user for a missing parameter instead of assuming one. When a tool call returns an error, the agent receives it and may retry. So the error alone does not count as a task failure. We also evaluate ten Single LLMs without tool access or live observations.

\textbf{Evaluation.}
We follow the UrbanEval protocol. TIC, IOA, and the cost derive directly from the execution traces; the remaining metrics require judging the content of an answer or trajectory rather than applying a rule. Qwen3-Max at temperature~$0$ scores them with system identities masked to limit bias~\cite{wang2023fair,zheng2023judge}, grading TSR and MIR against fixed per-query criteria and VPAR against a fixed trajectory rubric. To keep the benchmark neutral, we froze the UrbanEval queries and schema before finalizing UrbanAgent, and annotated the schema independently of its development. All systems query the same MCP services, and every raw return is logged. Because these responses vary over time, the schema grades time-varying facts against each run's own returns rather than a fixed key. So no system is credited or penalized for the particular values a service returned during its run.

\subsection{Main Results}
\begin{figure*}[t]
    \centering
    \includegraphics[width=\textwidth]{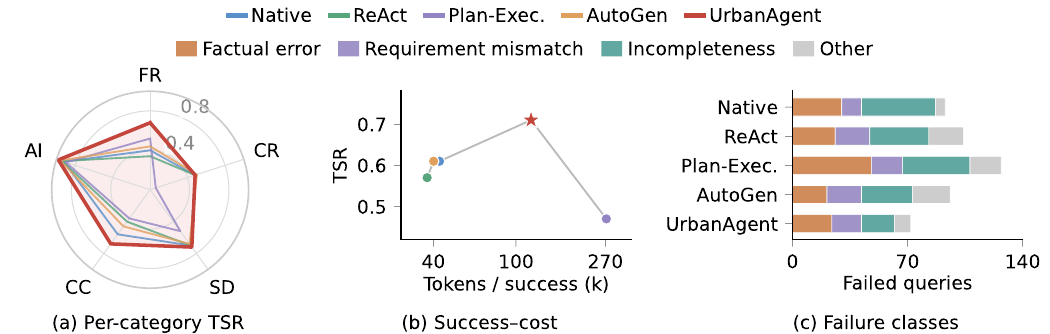}
    \caption{Main results and analysis on GPT-5-mini. (a)~TSR by query type: factual retrieval (FR), complex retrieval across independent services (CR), sequential dependency (SD), conditional or combinatorial decision-making (CC), and ambiguous or incomplete intent (AI). (b)~TSR vs tokens per successful task. (c)~Failure counts by class including factual error, requirement mismatch, incompleteness, and other over the 200 executable queries.}
    \label{fig:results}
\end{figure*}

\noindent\textbf{Controlled comparison (Q1).} Over all 250 queries UrbanAgent attains an overall TSR of \textbf{71\%}, against 61\% for the strongest baselines, Native tool-calling and AutoGen (Table~\ref{tab:main_results}). The 50 ambiguous or incomplete intent queries test clarification rather than execution, so we isolate the 200 executable queries separately: UrbanAgent completes 128 (64.0\%) against 107 (53.5\%) for Native tool-calling, a 10.5-point gain at matched model and tool access. The gain incurs a higher token cost. 118.3k per successful task is about three times the 37.3--42.8k of Native tool-calling, ReAct, and AutoGen.(Figure~\ref{fig:results}b). The higher token cost buys reliability on the hard requests. On the 200 executable queries UrbanAgent solves 21 more than Native tool-calling.

The margin concentrates in two categories (Figure~\ref{fig:results}a): UrbanAgent exceeds the strongest baseline by 16 points on factual retrieval (0.68 vs.\ 0.52) and 12 on conditional or combinatorial decision-making (0.68 vs.\ 0.56). It narrows on sequential dependency, where the leading systems fall between 0.68 and 0.72, and reverses on complex retrieval across independent services, where AutoGen edges UrbanAgent 0.50 to 0.48.

\textbf{A stronger base model does not bridge the gap.} The ten Single LLMs (Table~\ref{tab:single_llm}) include the strongest available models. But all fall within a narrow 0.22--0.26 overall TSR. Their successes are almost entirely clarification: 70--91\% arise on the ambiguous or incomplete intent category, while no executable category exceeds 0.22 and the best scores on the conditional or combinatorial decision-making and sequential dependency categories reach only 0.12. Without tools, even a frontier model flags a missing input but cannot satisfy a request that depends on live urban observations. Tool use closes this gap: on GPT-5-mini, TSR rises from 0.22 without tools to 0.71 with UrbanAgent, a 49-point difference. Scale alone does not close this gap. The newest models stay within the same narrow band as the older ones. The bottleneck is access to live data, not model capacity.

\begin{table}[!t]
\centering
\setlength{\tabcolsep}{5pt}
{\small
\begin{tabular}{l cccc}
\toprule
Model & TSR~$\uparrow$ & MIR~$\uparrow$ & Tok.~$\downarrow$ & T/Succ.~$\downarrow$ \\
\midrule
GPT-5-mini & 0.22 & \textbf{1.00} & 1.6 & 7.2 \\
Gemini-2.5-flash & \textbf{0.26} & 0.94 & 0.5 & 1.8 \\
DeepSeek-V4-flash & \underline{0.25} & 0.92 & 0.5 & 1.9 \\
Qwen3-235B-A22B & 0.24 & \underline{0.96} & \textbf{0.3} & \textbf{1.2} \\
GPT-5 & \textbf{0.26} & \textbf{1.00} & 2.4 & 9.2 \\
Claude-Opus-4.5 & \underline{0.25} & \textbf{1.00} & \underline{0.4} & \underline{1.7} \\
Gemini-3-Pro & \textbf{0.26} & 0.94 & 0.9 & 3.4 \\
DeepSeek-R1 & 0.22 & 0.88 & 0.6 & 2.8 \\
Kimi-K2.5 & \textbf{0.26} & 0.92 & 0.9 & 3.4 \\
GLM-5 & \underline{0.25} & 0.88 & 1.6 & 6.4 \\
\bottomrule
\end{tabular}
}
\caption{Results for ten single LLMs prompted directly, without tools or live observations. The process metrics therefore do not apply.}
\label{tab:single_llm}
\end{table}

\subsection{Analysis}
\noindent\textbf{Locating the advantage (Q2).} By separating results from process, UrbanEval also localizes the 10.5-point gap. On GPT-5-mini the tool-augmented systems already coincide on whether the right tools are called in the right order: TIC is 0.89 for UrbanAgent against 0.86 for the strongest baseline, and IOA 1.00 against 0.99. Clarification is equally saturated: MIR ranges from 0.90 to 0.98, and every system identifies the missing input and requests it. UrbanAgent departs only in never acting on an assumed value, returning no assumption across the 50 ambiguous or incomplete intent queries against 5--8 for the baselines. Coverage, ordering, and clarification thus do not explain the gap. There are two behaviors can explain including augmentation beyond the minimal toolchain and completion of constrained tasks.

\textbf{Proactive augmentation.} Among the process metrics, only augmentation separates the systems: VPAR reaches 0.96 for UrbanAgent against 0.84 for the strongest baseline. Valid augmentation takes four forms: precision refinement (narrowing a coarse result to an exact one), expanded retrieval (evidence gathered beyond the minimal toolchain), error recovery (a failed call retried with new arguments or another tool), and cross-validation (a fact confirmed from two independent calls). The judge credits at least one on 95 of the 99 Level-2 runs in which UrbanAgent issued a tool call; the remaining run issued none, leaving nothing to augment. Refinement appears in 85 trajectories and expanded retrieval in 54, error recovery in 8 and cross-validation in 10; the categories overlap, so a trajectory may bear several. UrbanAgent thus supplements and cross-checks steps that a minimal, correctly ordered toolchain omits. Refinement is the most common form. A coarse match becomes the exact value that the request needs.

\textbf{Task completeness.} The second locus is completion, which the failure classes expose. Each failed run receives one dominant class: a factual error (a claim its observations do not support), a requirement mismatch (a result that violates a stated constraint), task incompleteness (a required step left undone), or other. UrbanAgent fails 72 of the 200 executable queries, against 93 for Native tool-calling, 104 for ReAct, 127 for Plan-and-Execute, and 96 for AutoGen, and the classes diverge most on incompleteness (Figure~\ref{fig:results}c): UrbanAgent leaves 20 tasks incomplete against 31--45 for the baselines. One factual retrieval query illustrates the pattern: it requests hotels near Kunming South Railway Station with availability from 12--15 July 2026. All four baselines retrieve hotels but none verifies availability for those dates. Native tool-calling and AutoGen defer the check, ReAct returns no verified hotel, and Plan-and-Execute spends 25 LLM calls on candidates and routes yet never queries rooms. UrbanAgent checks several nearby hotels for the dates and reports that none has rooms, completing the query the baselines abandon. This case is typical of constrained retrieval. UrbanAgent tests each candidate before it answers.

\textbf{Robustness across base models (Q3).} UrbanAgent ranks first in TSR across all four base models and first in MIR within each tool-augmented block (Table~\ref{tab:main_results}; a few tool-free Single LLMs reach a higher MIR of 1.00, Table~\ref{tab:single_llm}). The rank is stable but the margin is not: +10 points on GPT-5-mini, +14 on Gemini-2.5-flash, and +9 on Qwen3-235B-A22B, yet +2 on DeepSeek-V4-flash. The margin tracks how strong the baselines already are. On DeepSeek-V4-flash the best baseline already reaches 0.50 TSR.

\noindent\textbf{Limitations.}
UrbanAgent is more token-intensive, and its gains vary across tasks and backbones; evaluation with time-varying services and an LLM judge does not establish generalization to unseen cities or domains.

\section{Conclusion}
We propose UrbanAgent, a tool-augmented agent that turns a natural-language urban request into an executable cross-system workflow. Meanwhile, we built UrbanEval, a 250-query benchmark that scores not only whether a request succeeds but how it is executed. Under GPT-5-mini, UrbanAgent reaches 71\% TSR, 10 points above the strongest baseline, and completes 64.0\% of executable requests against 53.5\%. Future work will isolate each component's contribution and add user studies and in-the-wild deployment.

{\small
\bibliography{urbanagent}
}

\end{document}